\documentclass[letterpaper]{article} % DO NOT CHANGE THIS
\usepackage[draft]{aaai2027}  % DO NOT CHANGE THIS
\usepackage[hyphens]{url}  % DO NOT CHANGE THIS
\usepackage{graphicx} % DO NOT CHANGE THIS
\usepackage{natbib}  % DO NOT CHANGE THIS AND DO NOT ADD ANY OPTIONS TO IT
\usepackage{caption} % DO NOT CHANGE THIS AND DO NOT ADD ANY OPTIONS TO IT
\usepackage{booktabs}

\usepackage{amsmath}
\usepackage{amssymb}
\usepackage{dsfont}   % provides \mathds{1} used in the method section
\usepackage{multirow} % provides \multirow used in the main results table
\usepackage{xcolor}   % provides \textcolor used for improvement annotations
\definecolor{ForestGreen}{RGB}{34,139,34} % improvement-annotation color (dvipsnames unavailable due to option clash)
\usepackage{placeins} % provides \FloatBarrier to flush experiment floats before the Conclusion

\providecommand{\Description}[1]{}

\title{Teaching MLLMs to Say No: Generalized Referring Expression Comprehension via Refusal Calibrated GRPO}

\author{
Xuzheng Yang,
Jun Ling,
Tao Huang,
Caiyan Qin,
Peng Wang
}

\affiliations{
School of Computer Science and Engineering,\\
University of Electronic Science and Technology of China\\
\texttt{yangxuzheng@std.uestc.edu.cn},
\texttt{p.wang6@hotmail.com}\\
}

\begin{document}

\maketitle

\begin{abstract}

We tackle the challenging yet underexplored task of Generalized Referring Expression Comprehension (GREC), which requires a model to localize the object described by a textual expression when it exists (positive sample) and to refuse output when it does not (negative sample). Although Multimodal Large Language Models (MLLMs) excel at localizing existing objects, they often fail to reject nonexistent ones due to the absence of negative samples during training, producing hallucinated bounding boxes. Existing post-training approaches such as supervised fine-tuning (SFT) and reinforcement learning (RL) enhance refusal behavior but usually degrade localization accuracy on positive samples, undermining the model’s core competence.
To address this, we propose Refusal-Calibrated Group Relative Policy Optimization (RC-GRPO), a calibrated RL strategy that strengthens the refusal ability of MLLMs while preserving localization performance. It enforces ``None'' outputs in rollouts for valid advantage estimation on negative samples and applies a penalty to prevent over-refusal on positives, achieving a balanced trade-off between accuracy and reliability. A second-stage reasoning reinforcement further consolidates causal understanding and interpretability. Experiments on three GREC benchmarks demonstrate that RC-GRPO attains superior localization accuracy while maintaining strong refusal capability.
\end{abstract}

\section{Introduction}
\label{sec:intro}

\begin{figure}[t]
  \centering
  \includegraphics[width=0.9\columnwidth]{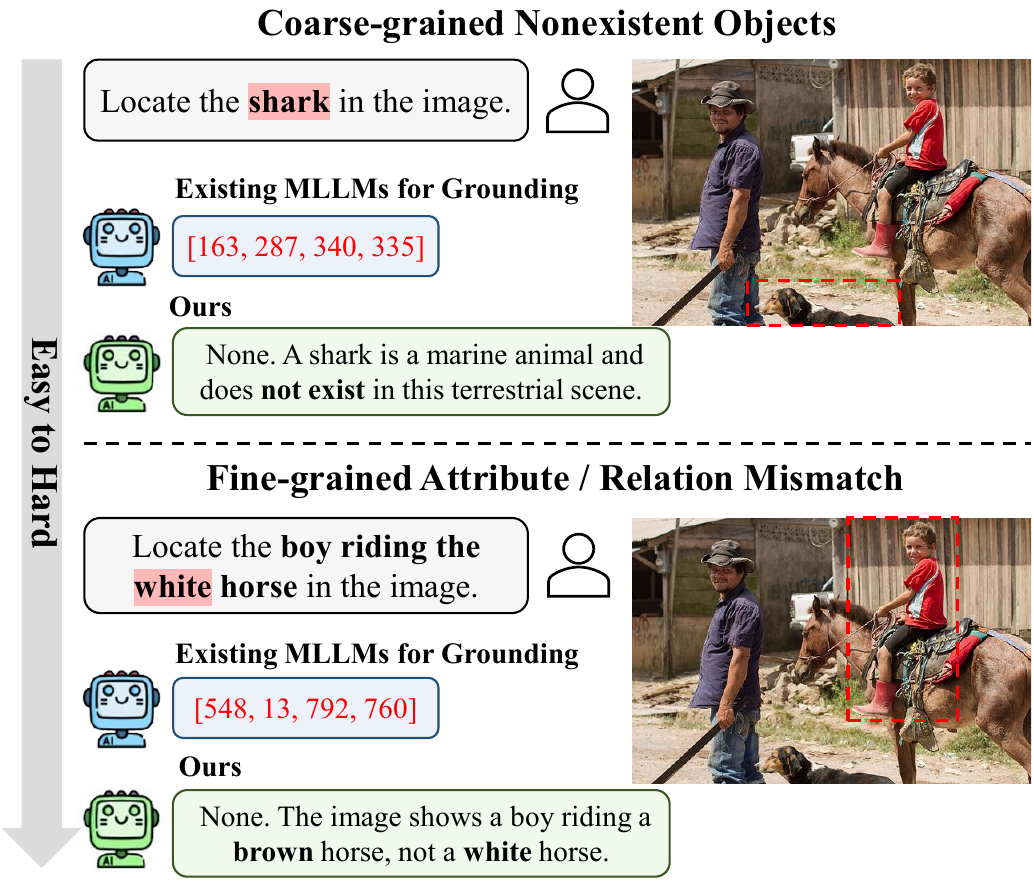}
  \caption{\textbf{Existing MLLMs fail to reject nonexistent cases.} In the upper example, when the referred object is absent, Qwen2.5-VL-32B~\cite{Qwen2.5-VL} incorrectly localizes an unrelated region. In the lower example, Qwen3-VL-30B~\cite{qwen3} overlooks an attribute mismatch and still outputs a bounding box. In contrast, our method successfully identifies the absence of the referred target and predicts a refusal with clear, interpretable reasoning.}
  \label{reject_case}
\end{figure}
Referring Expression Comprehension (REC) is a fundamental vision-language task that localizes a target object in an image from a given textual expression~\cite{mao2016generation,yu2016modeling,nagaraja2016modeling}. With broad applications in visual grounding, human–robot interaction, and embodied AI, it has attracted extensive research attention. However, existing REC formulations implicitly assume the described object always exists—an assumption that often fails in real-world settings, e.g., when an object moves out of a robot’s field of view, where the model should instead recognize its absence and refuse to localize. To address this, the Generalized Referring Expression Comprehension (GREC)~\cite{he2023grec} task extends REC to handle both positive samples (object present) and negative samples (object absent). Despite its practical significance, GREC remains insufficiently explored, leaving open the challenge of building models that both accurately localize existing objects and reliably reject nonexistent ones.

Thanks to their strong perceptual and cross-modal reasoning abilities, Multimodal Large Language Models (MLLMs) achieve impressive performance on REC tasks~\cite{Qwen2.5-VL,wang2025internvl3_5,chen2023shikra,li2024groundinggpt,wang2023cogvlm,qi2024cogcom}, yet still struggle to refuse localization for nonexistent cases. As shown in Figure~\ref{reject_case}, when the referred object is absent or its attributes or relations mismatch the image, existing grounding MLLMs fail to reject these cases and instead generate hallucinated bounding boxes.
This limitation stems from the absence of negative samples during large-scale pre-training. A natural solution is post-training with both positive and negative samples to explicitly encourage refusal. The central challenge lies in enhancing rejection ability without compromising localization accuracy on positive samples—a trade-off that remains unresolved. 

Conventional post-training approaches, such as supervised fine-tuning (SFT) and RL-based methods like DPO~\cite{rafailov2023direct} or GRPO~\cite{shao2024deepseekmath}, can substantially improve refusal on negative samples. Yet this gain often comes at the cost of a drastic decline in localization accuracy on positive samples, rendering the models unusable for practical REC applications.

To overcome this, we propose Refusal-Calibrated Group Relative Policy Optimization (RC-GRPO), a reinforcement learning–based post-training framework. It builds upon GRPO, which estimates relative advantages across batch rollouts without an explicit reward model. However, standard GRPO fails under GREC because MLLMs lack the inherent ability to refuse on negative samples—typically hallucinating bounding boxes instead—so the model obtains no valid advantages or gradient signals to strengthen rejection. To resolve this, we introduce a refusal calibration mechanism that enforces ``None'' outputs during rollout, ensuring valid advantage computation and gradient feedback for rejection learning. Since this can cause over-refusal that degrades localization on positives, we add a positive-sample penalty that discourages unnecessary ``None'' predictions and preserves core localization capacity. This adversarially calibrated optimization achieves a robust localization–refusal equilibrium that conventional post-training strategies fail to reach.

Although outputting ``None'' is straightforward, it does not necessarily indicate that the model truly understands the referred object’s absence. We therefore reinforce its causal understanding by encouraging concise explanations for refusal decisions. This reinforcement consolidates refusal capacity—ensuring ``None'' predictions stem from genuine comprehension rather than superficial patterns—while improving interpretability. On three challenging GREC benchmarks, our method achieves a markedly better balance between positive-sample localization and negative-sample refusal reliability, making GREC substantially more practical for real-world deployment.

The main contributions of this work are as follows:

\begin{itemize}

\item We study the underexplored Generalized Referring Expression Comprehension (GREC) task, emphasizing the importance of balancing localization accuracy and refusal reliability for practical deployment.

\item We propose Refusal-Calibrated Group Relative Policy Optimization (RC-GRPO), a reinforcement learning–based post-training framework that introduces a refusal calibration mechanism and a positive-sample penalty to jointly optimize rejection and localization.

\item We introduce a reason-generation stage that reinforces the model’s causal understanding of object absence, grounding refusals in genuine semantic reasoning while improving interpretability.

\item Extensive experiments on three GREC benchmarks demonstrate that our approach achieves a superior trade-off between positive and negative samples over conventional post-training strategies.

\end{itemize}

\section{Related Work}
\label{sec:related}

\textbf{Generalized Referring Expression Comprehension.}
Referring Expression Comprehension (REC) requires models to localize a target object in an image based on a natural language description. Early benchmarks such as RefCOCO, RefCOCO+, and RefCOCOg~\cite{mao2016generation,yu2016modeling,nagaraja2016modeling} established foundational evaluation protocols. These datasets are relatively simple, letting models rely on shallow keyword correlations or dominant visual cues. 
% Moreover, they assume that each referring expression corresponds to an existing object, overlooking more complex and realistic scenarios where the referred object may be absent.
They assume targets always exist, overlooking realistic scenarios where objects may be absent.

Generalized Referring Expression Comprehension (GREC)~\cite{he2023grec} extends REC by allowing referring expressions to correspond to multiple objects or none at all, reflecting realistic open-world scenarios. Recent studies~\cite{kurita2023refego,schulter2023omnilabel,yao2024evaluate,Xie2023DescribedOD,liu2024finecops} emphasize modeling nonexistent-object cases in REC. 
% For example, RefEgo~\cite{kurita2023refego}, a video-based REC benchmark, treats out-of-frame objects as nonexistent instances. OmniLabel~\cite{schulter2023omnilabel} and OVDEval~\cite{yao2024evaluate} address situations where a referring expression may match multiple or zero targets, focusing on open-vocabulary detection. 
Specifically, RefEgo~\cite{kurita2023refego} treats out-of-frame objects as nonexistent, while OmniLabel~\cite{schulter2023omnilabel} and OVDEval~\cite{yao2024evaluate} address multi-target or zero-target scenarios.
D\textsuperscript{3}~\cite{Xie2023DescribedOD} pairs each expression with several unrelated images from the same context, while FineCops-Ref~\cite{liu2024finecops} introduces hard negatives through fine-grained textual perturbations and image edits. Collectively, these efforts highlight that handling nonexistent objects has become a central, unresolved challenge for Multimodal Large Language Models (MLLMs) in real-world applications.
%Generalized Referring Expression Comprehension (GREC)~\cite{he2023grec} aims to overcome the limitations of traditional REC by extending the task to multiple targets or no specific target, reflecting more realistic scenarios. Recent researches~\cite{kurita2023refego,schulter2023omnilabel,yao2024evaluate,Xie2023DescribedOD,liu2024finecops} have increasingly recognized the importance of nonexistent objects in REC. For instance, RefEgo~\cite{kurita2023refego}, a video-based REC benchmark, treats a referred object that becomes out-of-frame as a nonexistent case. OmniLabel~\cite{schulter2023omnilabel} and OVDEval~\cite{yao2024evaluate} focus on cases where there may be multiple or no objects, emphasizing open-vocabulary detection. D\textsuperscript{3}~\cite{Xie2023DescribedOD} generates negative pairs by selecting multiple negative images for each expression from the same scenarios, thereby representing nonexistent cases. FineCops-Ref~\cite{liu2024finecops} introduces hard negatives by applying fine-grained text perturbations and image edits to positive samples. The handling of nonexistent objects has emerged as a key challenge for MLLMs in real-world applications.

\textbf{Existing Efforts for Nonexistent Rejection.}
Early studies~\cite{yan2023universal,kamath2021mdetr,luo2020multi,ding2021vision} addressed the nonexistent-object problem by applying confidence thresholds to suppress low-scoring bounding boxes. Subsequent research decouples existence prediction from localization, with several works~\cite{hemanthage2024recantformer,wang2025hierarchical,dai2025improving,yu2024revisiting} introducing dedicated network heads for existence discrimination. 
% For example, RECANTFormer~\cite{hemanthage2024recantformer} incorporates a validity-prediction head to filter irrelevant candidate boxes, while HieA2G~\cite{wang2025hierarchical} employs an adaptive grounding counter to estimate the number of referred objects. InstanceVG~\cite{dai2025improving} performs binary classification to determine whether a target exists. 
For instance, RECANTFormer~\cite{hemanthage2024recantformer}, HieA2G~\cite{wang2025hierarchical}, and InstanceVG~\cite{dai2025improving} introduce specialized heads for validity prediction or object counting.
Although effective at rejecting nonexistent cases, these designs are closely tied to specific architectures, limiting their scalability and applicability to general-purpose MLLMs.

In contrast, efforts to equip MLLMs with rejection capability remain limited. Representative approaches such as ROD-MLLM~\cite{yin2025rod} and CRS~\cite{yang2025new} adopt a two-stage paradigm: a detector proposes candidate bounding boxes, then an MLLM evaluates each candidate's alignment with the referring expression to filter out invalid regions. While effective, these methods add computational overhead and fail to address the intrinsic refusal deficiency of MLLMs, which still generate spurious bounding boxes when directly prompted to locate a nonexistent object.
% Our work differs fundamentally by approaching the problem from a training perspective. Instead of relying on external detectors or architectural modifications, we develop a reinforcement-learning-based post-training framework that intrinsically enables MLLMs to ``say no''—empowering them to reject nonexistent objects reliably and with grounded reasoning.
Unlike these modular or architectural approaches, our RL-based post-training framework enables MLLMs to intrinsically "say no" with grounded reasoning.

\textbf{Multimodal Hallucination.}
Multimodal hallucination in MLLMs is fundamentally an issue of image-text misalignment, where models fabricate visual entities that do not exist. Extensive research mitigates it through robust instruction tuning~\cite{liu2023mitigating}, dynamic decoding interventions~\cite{wang2024mllm}, and post-hoc corrections~\cite{yang2025nullu}, generally aligning free-form text outputs with visual context in descriptive or generative tasks (e.g., captioning or VQA).

Our work addresses a severe manifestation of hallucination in visual grounding: the failure to reject nonexistent objects under instruction-observation mismatches. When targets are absent or occluded, MLLMs, driven by strong textual priors~\cite{fu2025hidden, tong2024eyes}, succumb to compulsory grounding prompts and hallucinate coordinates instead of refusing, risking erroneous robotic actions. Existing mitigations, effective for text generation, struggle on dense coordinate regression, where hallucination suppression must be balanced against foundational grounding. Unlike prior post-hoc or architectural fixes, our RL-based post-training internalizes causal refusal logic, preserving localization while enabling reliable rejection.
\section{Method}
\label{sec:method}

\begin{figure*}[t]
  \centering
  \includegraphics[width=0.9\textwidth]{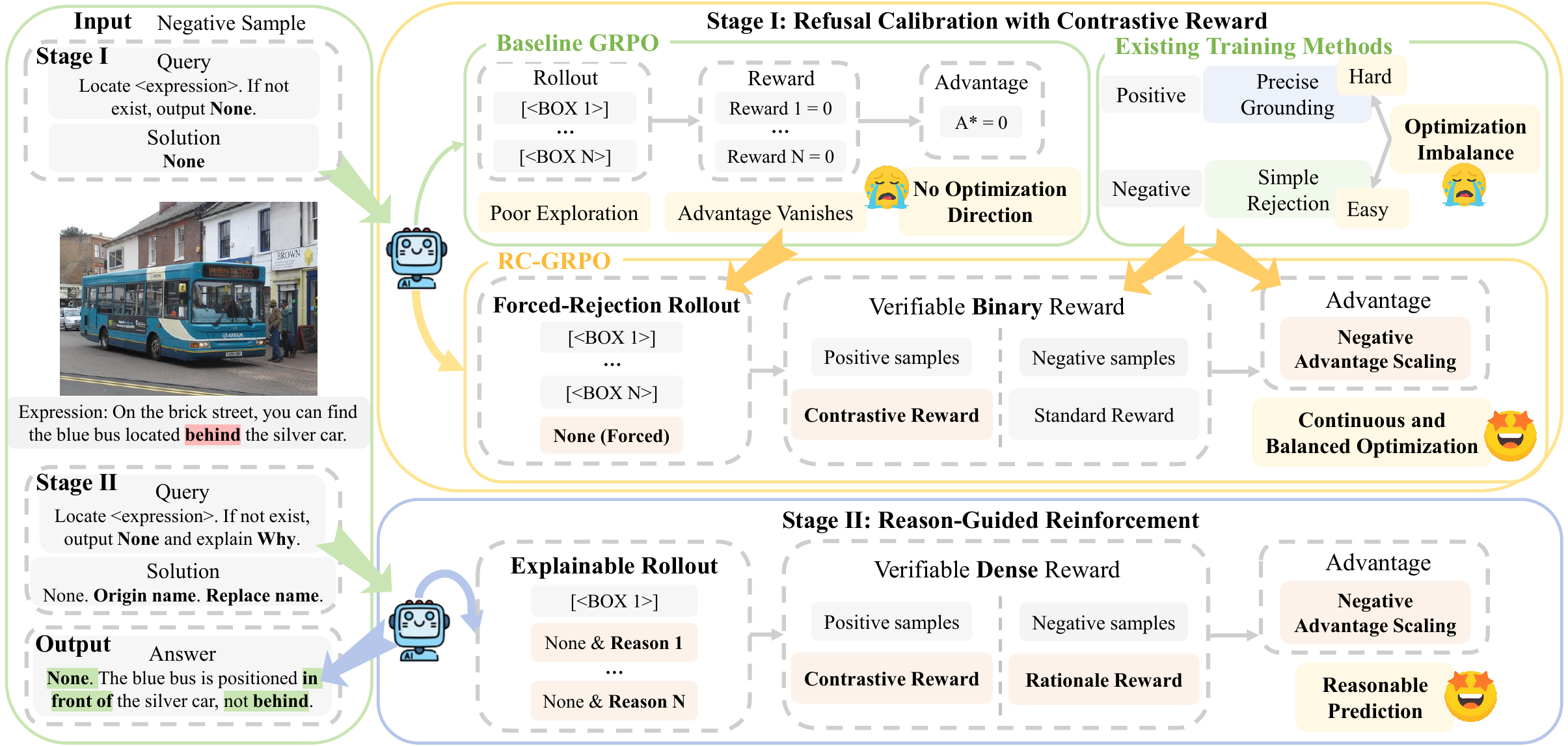}
  \caption{\textbf{The pipeline of RC-GRPO.} The left side shows examples of model input and output. Stage I (Refusal Calibration with Contrastive Reward) equips the model with balanced rejection ability by enforcing ``None'' outputs and applying contrastive rewards to preserve localization on positives. Stage II (Reason-Guided Reinforcement) further strengthens this behavior by encouraging the model to reason about the cause of absence, yielding more reliable and interpretable refusals.}
  \label{pipeline}
\end{figure*}

\subsection{Limitations of Existing Post-Training Methods}
\label{limitations}

\textbf{Lack of Rejection Training.}
In Referring Expression Comprehension (REC), conventional training strategies~\cite{Qwen2.5-VL,wang2025internvl3_5,chen2023shikra,li2024groundinggpt,wang2023cogvlm,qi2024cogcom} focus solely on optimizing localization accuracy, neglecting the critical nonexistent-object cases frequent in real-world deployments.
The loss function implicitly encourages the model to always output a bounding box—even a random prediction might occasionally overlap the target, whereas predicting ``None'' is consistently penalized. Consequently, as illustrated in Figure~\ref{reject_case}, grounding MLLMs fail to reject even obvious nonexistent cases; more capable models detect gross mismatches but still struggle with subtle attribute- or relation-level inconsistencies.
This observation raises a key question: given that standard training implicitly enforces compulsory bounding-box outputs, \textit{can simply including negative samples in training truly endow models with reliable rejection ability?}

\textbf{Optimization Imbalance Harms Positive Performance.}
To answer this, we constructed a mixed dataset with both positive and negative samples and evaluated mainstream training paradigms. Although each method improved rejection ability, all significantly degraded positive-sample performance, marked by a sharp increase in false negatives.
This degradation stems from an inherent optimization imbalance: positive samples require predicting and precisely localizing a bounding box, while negatives merely need a ``None'' output. This asymmetry in difficulty and optimization space biases the model toward the simpler ``None,'' causing over-refusal and substantial loss of localization accuracy.
Such training only teaches the model to say ``no,'' but not when or why. Since accurate localization on positive cases is REC's core competence, minimizing its degradation is crucial: we need a learning scheme that rejects nonexistent objects while maintaining strong positive-case performance—a balanced trade-off.

\textbf{SFT and DPO are Inherently Unsuitable for Grounding.}
Conventional post-training methods such as Supervised Fine-Tuning (SFT) and Direct Preference Optimization (DPO)~\cite{rafailov2023direct} compute token-level losses between model outputs and reference annotations. However, in grounding, outputs are serialized bounding-box coordinates, and token-level alignment fails to reflect spatial equivalence—slightly different coordinate tokens may map to nearly identical regions. Such rigid supervision over-penalizes semantically correct but textually different predictions, limiting robust spatial grounding.

\textbf{GRPO Struggles in Early Exploration.}
We also evaluated Group Relative Policy Optimization (GRPO)~\cite{shao2024deepseekmath} and observed poor exploration in early training. Since the base model lacks inherent rejection capability, even prompts explicitly instructing it to output ``None'' for nonexistent cases fail to elicit the correct response. Negative-sample rollouts thus predominantly produce bounding boxes, yielding zero rewards across the batch and vanishing advantage estimates; without meaningful gradient signals, learning quickly stalls.
Previous work~\cite{guo2025deepseek,huang2025vision,bai2025univg,yang2025r1} mitigates this with SFT as a cold start, but as discussed, SFT introduces severe bias that catastrophically harms positive-sample accuracy, making it unsuitable here.

\subsection{Refusal-Calibrated Group Relative Policy Optimization}
To address the limitations discussed above, we propose a two-stage reinforcement learning framework named Refusal-Calibrated Group Relative Policy Optimization (RC-GRPO), illustrated in Figure~\ref{pipeline}.
Here, ``calibration'' refers to correcting the model’s inherent bias—its tendency to compulsively predict bounding boxes—rather than traditional probabilistic calibration.
Stage I (Refusal Calibration with Contrastive Reward) establishes the model's basic rejection capability while maintaining localization accuracy.
Stage II (Reason-Guided Reinforcement) further strengthens the model’s causal understanding of absence by encouraging it to generate reasoning for its refusals.
%To address the above issues, we propose a two-stage training strategy named Refusal-Calibrated Group Relative Policy Optimization (RC-GRPO), as illustrated in Figure~\ref{pipeline}. The first stage, Forced-Rejection Learning, focuses on establishing the model’s basic rejection ability. The second stage, Explainable Refinement, further enables the model to provide reliable explanations while rejecting.

\subsubsection{Training Data}
\label{training_data}
We adopt FineCops-Ref~\cite{liu2024finecops}, a fine-grained and challenging benchmark supporting controllable difficulty and multi-hop reasoning, which constructs hard negative samples via linguistic perturbations and image editing.
From its training split, we select 2,000 high-quality samples across multiple difficulty levels, balanced between 1,000 positive and 1,000 negative instances. The same data are used in both stages, differing only in prompting.
In Stage I, the instruction ``If the object does not exist, output: There are none.'' is appended, and the target output for negatives is ``There are none.''
In Stage II, we extend the prompt with ``Also briefly explain the reason why it does not exist.'' The target output still includes ``There are none.'' but now incorporates annotated construction information for rule-based evaluation of the generated explanations.
%FineCops-Ref~\cite{liu2024finecops} is a fine-grained and challenging dataset that supports controllable difficulty and multi-hop reasoning. It constructs hard negative samples through linguistic perturbations and image editing, and annotates the negative construction process with fields such as ``negative\_type'', ``origin\_name'', and ``replace\_name''. We select 2,000 high-quality samples from its training set, covering different difficulty levels, balanced with 1,000 positive and 1,000 negative samples. The same data is used across both stages. The only difference is in prompting: in Stage 1, we append the instruction ``If the object does not exist, output: There are none.'', and for negative samples the target output is ``There are none.''. In Stage 2, we additionally include ``Also briefly explain the reason why it does not exist.''. The target output for negative samples remains ``There are none.'', but now includes the annotated construction information to facilitate rule-based evaluation of the generated explanation.

\subsubsection{Refusal Calibration with Contrastive Reward}
The first stage equips the model with fundamental rejection ability while minimizing positive-sample degradation—a transition from 0 to 1—via three key components:

\textbf{Forced-Rejection Rollout.}
Since the base model rarely explores ``None'' during early training, valid advantages cannot be computed for negatives when all rollouts produce bounding boxes. We thus enforce rejection during rollout: if no completion contains ``None,'' we apply constrained beam search to force the output ``There are none.'' This ensures the rollout group includes valid refusals, enabling meaningful advantage estimation for both positive and negative samples.
Unlike prior methods~\cite{yan2025learning,cui2025process,zhang2025critic} that rely on stronger external models for diverse exploration, our approach is self-contained—the policy model itself produces ``None'' and joins gradient updates, internalizing refusal and stabilizing early training.

\textbf{Contrastive Reward for Positive Samples.}
To encourage correct localization while avoiding over-refusal, we introduce a contrastive reward:
\begin{equation}
R_{\text{pos}} = 
\begin{cases} 
1, & \text{if IoU} \ge 0.5, \\ 
0, & \text{if IoU} < 0.5, \\ 
-1, & \text{if the model predicts ``None''}.
\end{cases}
\end{equation}
\begin{equation}
R_{\text{neg}} = 
\begin{cases} 
1, & \text{if the model predicts ``None''}, \\ 
0, & \text{otherwise}. 
\end{cases}
\end{equation}
As defined above, this design teaches the model that even a poor bounding box is preferable to refusing when an object exists (hence the $-1$ penalty for a false ``None''), reducing false rejections, while negatives are rewarded only for a correct ``None.'' Thus every sample—positive or negative—contributes a meaningful advantage signal, stabilizing learning and accelerating convergence.

\textbf{Negative Advantage Scaling.}
Because rejecting negatives is inherently easier than localizing positives, the model may overfit to “None.” To mitigate this imbalance, we scale the advantages of negative samples by a coefficient \( \alpha \in (0,1] \):
%As discussed above, there exists a substantial imbalance in optimization between positive and negative samples. Saying ``no'' for negative samples is inherently simpler, which can lead the model to overfit to the rejection behavior and degrade positive-sample accuracy.
%To mitigate this imbalance, we introduce negative advantage scaling, inspired by prior work~\cite{su2025learning}.
%Specifically, the advantage of negative samples is scaled by a coefficient \( \alpha \in (0,1] \), formulated as:
\begin{equation}
A^* =
\begin{cases} 
A^*, & \text{for positive samples}, \\
\alpha \cdot A^*, & \text{for negative samples}.
\end{cases}
\end{equation}
This adjustment balances the optimization dynamics, preventing the model from overemphasizing negative-sample learning while maintaining stable policy updates.

\subsubsection{Reason-Guided Reinforcement}
To ensure that the model's refusal stems from genuine semantic comprehension rather than superficial pattern matching, we introduce a Rationale Reward $r_{\text{reason}}$ in the second stage that rewards the model for explaining the specific expression–image mismatch.

% \textbf{Formalizing Semantic Patterns.} Each negative sample is synthesized from a positive triplet $(\text{Image } I, \text{Expression } E, \text{BBox } B)$ by applying a linguistic perturbation to a specific semantic component $P$ (e.g., a relation or attribute). This process defines two distinct patterns:
% \begin{itemize}
%     \item \textbf{Incorrect Pattern ($P_{\text{inc}}$):} The misleading or ``replaced'' cue introduced into the expression that contradicts the image (e.g., changing ``in front of'' to ``\textbf{behind}'').
%     \item \textbf{Original Pattern ($P_{\text{orig}}$):} The genuine semantic component that accurately describes the image (e.g., the ground-truth relation ``\textbf{in front of}'').
% \end{itemize}
\textbf{Formalizing Semantic Patterns.} Each negative sample is synthesized from a positive triplet $(\text{Image } I, \text{Expression } E, \text{BBox } B)$ by perturbing a semantic component $P$ (e.g., a relation or attribute), defining two distinct patterns: the Incorrect Pattern ($P_{\text{inc}}$), a misleading cue mismatching the image (e.g., changing ``in front of'' to ``behind''), and the Original Pattern ($P_{\text{orig}}$), the genuine component that accurately describes the image (e.g., ``in front of'').

\textbf{Rationale Reward Formulation.} When the model predicts ``None'', we evaluate the generated explanation $S$ using a rule-based verification derived from the construction logic. The reasoning reward $r_{\text{reason}}$ is formally defined as:
\begin{equation}
r_{\text{reason}} = w_{\text{inc}} \cdot \mathds{1}(P_{\text{inc}} \in S) + w_{\text{orig}} \cdot \mathds{1}(P_{\text{orig}} \in S),
\end{equation}
where $\mathds{1}(\cdot)$ is the indicator function. Following the construction logic on FineCops-Ref, we set $w_{\text{inc}} = 0.5$ and $w_{\text{orig}} = 0.3$, resulting in a maximum reasoning reward of $0.8$ when both patterns are correctly identified. The total reward for negative samples $R_{\text{neg}}$ is then defined as:
\begin{equation}
R_{\text{neg}} =
\begin{cases}
r_{\text{reject}} + r_{\text{reason}}, & \text{if the model predicts ``None''}, \\
0, & \text{otherwise},
\end{cases}
\end{equation}
where $r_{\text{reject}}$ is a constant base reward (0.2) for a correct refusal. This additive structure provides dense, fine-grained feedback, incentivizing automated, reproducible identification of causal rejection—not just that a referent is missing, but why it is mismatched by specific attributes or relations.

To maintain balance, we also add a dense IoU-based reward for positive samples: when IoU $<$ 0.5, the reward equals the IoU value itself, encouraging gradual localization improvement, while Stage I's $\alpha$-scaling is applied to negative-sample advantages for stable joint optimization.
%Since the rationale reward introduces a denser feedback signal for negative samples, we symmetrically design a dense reward for positive samples to maintain balanced optimization.
%Specifically, when the IoU is below 0.5, the reward is set equal to the IoU value itself, thereby encouraging gradual improvement in localization.
%We continue to apply the $\alpha$-scaling strategy to negative-sample advantages to ensure stable and balanced optimization.

Overall, RC-GRPO combines IoU-based rewards with contrastive and rationale-driven reinforcement, balancing optimization across positive and negative samples while grounding refusals in causal reasoning.

\section{Experiment}

\begin{table*}[t]
\centering
% \caption{\textbf{Comparison with state-of-the-art Specialist and MLLM methods on three benchmarks.} Pr represents Pr@(F1 = 1, IoU $\ge$ 0.5), which is the overall precision when the IoU threshold is set to $\ge$ 0.5 and the F1 score of the predicted box equals 1. N-acc represents the accuracy of rejecting nonexistent referent expression.}
\caption{Comparison with state-of-the-art Specialist and MLLM methods on three benchmarks. Pr denotes Precision@(F1 = 1, IoU $\ge$ 0.5), a holistic metric over positive and negative samples, and a higher Pr indicates a strictly superior model. N-acc represents the accuracy of rejecting nonexistent referring expressions. Green: absolute gain over the base model.}

\resizebox{0.9\textwidth}{!}{

\begin{tabular}{l|cc|cccccc|ccc}
\toprule
\quad \multirow{3}{*}{\textbf{Method}} & \multicolumn{2}{c|}{\textbf{FineCops-Ref}} & \multicolumn{6}{c|}{\textbf{gRefCOCO}} & \multicolumn{3}{c}{\textbf{D\textsuperscript{3}}} \\
& \multicolumn{2}{c|}{test} & \multicolumn{2}{c}{val} & \multicolumn{2}{c}{test-A} & \multicolumn{2}{c|}{test-B} & Full & PRES & ABS \\
\multicolumn{1}{c|}{} & Pr. & \multicolumn{1}{c|}{N-acc.} & Pr. & N-acc. & Pr. & N-acc. & Pr. & \multicolumn{1}{c|}{N-acc.} & \multicolumn{3}{c}{AP} \\ 
\midrule
\multicolumn{10}{l}{\textit{\textbf{Specialist methods}}} \\
\midrule
% Grounding DINO & - & - & - & - & - & - & - & - & 20.7 & 20.1 & 22.5 \\
% MDETR & - & - & 42.7 & 36.3 & 50.0 & 34.5 & 36.5 & 31.0 & - & - & - \\
\quad UNINEXT-R50~\cite{yan2023universal} & - & - & 58.2 & 50.6 & 46.4 & 49.3 & 42.9 & 48.2 & - & - & - \\
\quad HieA2G-R101~\cite{wang2025hierarchical} & - & - & 67.8 & 60.3 & 66.0 & 60.1 & 56.5 & 56.0 & - & - & - \\
\quad InstanceVG~\cite{dai2025improving} & - & - & 73.5 & 72.8 & 70.2 & 71.1 & 60.8 & 65.2 & - & - & - \\
\midrule
\multicolumn{10}{l}{\textit{\textbf{MLLM methods}}} \\ 
\midrule
% \multicolumn{10}{l}{\textit{{\color{gray} MLLM baseline}}} \\ 
\quad Ferret-7B~\cite{you2024ferret} & - & - & 54.8 & 48.9 & 49.5 & 45.2 & 43.5 & 43.8 & - & - & - \\
\quad Qwen2.5-VL-7B-Instruct~\cite{Qwen2.5-VL} & 56.4 & 5.3 & 38.2 & 6.4 & 47.2 & 3.4 & 38.3 & 4.2 & 26.4 & 26.3 & 26.7 \\
\quad Qwen3-VL-4B-Instruct~\cite{qwen3}   & 63.5 & 58.3 & 72.4 & 77.2 & 68.0 & 70.2 & 62.2 & 70.2 & 34.8 & 36.6 & 29.3 \\
% \cmidrule(lr){1-12}
% \multicolumn{10}{l}{\textit{{\color{gray} MLLM with a larger parameter scale}}} \\ 
\quad Qwen3-VL-235B-A22B-Instruct~\cite{qwen3} & 69.9 & 51.6 & 59.5 & 65.7 & 58.3 & 66.4 & 52.3 & 56.5 & 36.2 & 37.4 & 32.5 \\
% \cmidrule(lr){1-12}
% \multicolumn{10}{l}{\textit{{\color{gray} Previous SOTA methods}}} \\ 
\quad ROD-MLLM(Vicuna-7B)~\cite{yin2025rod} & - & - & 53.2 & 51.8 & 55.4 & 63.7 & 51.8 & 57.9 & 29.7 & 30.0 & 28.7 \\
\quad CRS(Qwen2.5-7B)~\cite{yang2025new} & 66.5 & 79.5 & 54.9 & 85.6 & 45.9 & 84.3 & 45.6 & 75.9 & 34.8 & 37.0 & 28.3 \\
\quad CRS(Qwen3-4B)~\cite{yang2025new}   & 65.0 & 83.9 & 56.3 & 89.5 & 46.4 & 87.3 & 47.5 & 84.0 & 36.1 & \textbf{39.0} & 27.2 \\
\midrule
\multicolumn{10}{l}{\textit{\textbf{Ours}}} \\
\midrule
\quad RC-GRPO-I(Qwen2.5-7B)   & \shortstack{66.6\\{\scriptsize\color{ForestGreen}$\uparrow$10.2}} & \shortstack{59.5\\{\scriptsize\color{ForestGreen}$\uparrow$54.2}} & \shortstack{62.8\\{\scriptsize\color{ForestGreen}$\uparrow$24.6}} & \shortstack{67.6\\{\scriptsize\color{ForestGreen}$\uparrow$61.2}} & \shortstack{59.9\\{\scriptsize\color{ForestGreen}$\uparrow$12.7}} & \shortstack{61.0\\{\scriptsize\color{ForestGreen}$\uparrow$57.6}} & \shortstack{52.6\\{\scriptsize\color{ForestGreen}$\uparrow$14.3}} & \shortstack{57.1\\{\scriptsize\color{ForestGreen}$\uparrow$52.9}} & \shortstack{33.8\\{\scriptsize\color{ForestGreen}$\uparrow$7.4}} & \shortstack{35.3\\{\scriptsize\color{ForestGreen}$\uparrow$9.0}} & \shortstack{29.2\\{\scriptsize\color{ForestGreen}$\uparrow$2.5}} \\
\quad RC-GRPO-II(Qwen2.5-7B)  & \shortstack{67.4\\{\scriptsize\color{ForestGreen}$\uparrow$11.0}} & \shortstack{61.1\\{\scriptsize\color{ForestGreen}$\uparrow$55.8}} & \shortstack{64.9\\{\scriptsize\color{ForestGreen}$\uparrow$26.7}} & \shortstack{71.5\\{\scriptsize\color{ForestGreen}$\uparrow$65.1}} & \shortstack{62.1\\{\scriptsize\color{ForestGreen}$\uparrow$14.9}} & \shortstack{66.2\\{\scriptsize\color{ForestGreen}$\uparrow$62.8}} & \shortstack{54.5\\{\scriptsize\color{ForestGreen}$\uparrow$16.2}} & \shortstack{62.3\\{\scriptsize\color{ForestGreen}$\uparrow$58.1}} & \shortstack{34.0\\{\scriptsize\color{ForestGreen}$\uparrow$7.6}} & \shortstack{35.6\\{\scriptsize\color{ForestGreen}$\uparrow$9.3}} & \shortstack{29.2\\{\scriptsize\color{ForestGreen}$\uparrow$2.5}} \\
\quad RC-GRPO-I(Qwen3-4B)     & \shortstack{67.6\\{\scriptsize\color{ForestGreen}$\uparrow$4.1}} & \shortstack{69.3\\{\scriptsize\color{ForestGreen}$\uparrow$11.0}} & \shortstack{74.4\\{\scriptsize\color{ForestGreen}$\uparrow$2.0}} & \shortstack{84.7\\{\scriptsize\color{ForestGreen}$\uparrow$7.5}} & \shortstack{69.0\\{\scriptsize\color{ForestGreen}$\uparrow$1.0}} & \shortstack{81.3\\{\scriptsize\color{ForestGreen}$\uparrow$11.1}} & \shortstack{63.2\\{\scriptsize\color{ForestGreen}$\uparrow$1.0}} & \shortstack{77.5\\{\scriptsize\color{ForestGreen}$\uparrow$7.3}} & \shortstack{\textbf{37.3}\\{\scriptsize\color{ForestGreen}$\uparrow$2.5}} & \shortstack{38.4\\{\scriptsize\color{ForestGreen}$\uparrow$1.8}} & \shortstack{\textbf{33.8}\\{\scriptsize\color{ForestGreen}$\uparrow$4.5}} \\
\quad RC-GRPO-II(Qwen3-4B)    & \shortstack{\textbf{67.8}\\{\scriptsize\color{ForestGreen}$\uparrow$4.3}} & \shortstack{71.3\\{\scriptsize\color{ForestGreen}$\uparrow$13.0}} & \shortstack{\textbf{75.7}\\{\scriptsize\color{ForestGreen}$\uparrow$3.3}} & \shortstack{85.3\\{\scriptsize\color{ForestGreen}$\uparrow$8.1}} & \shortstack{\textbf{71.4}\\{\scriptsize\color{ForestGreen}$\uparrow$3.4}} & \shortstack{83.1\\{\scriptsize\color{ForestGreen}$\uparrow$12.9}} & \shortstack{\textbf{64.8}\\{\scriptsize\color{ForestGreen}$\uparrow$2.6}} & \shortstack{78.4\\{\scriptsize\color{ForestGreen}$\uparrow$8.2}} & \shortstack{36.8\\{\scriptsize\color{ForestGreen}$\uparrow$2.0}} & \shortstack{37.8\\{\scriptsize\color{ForestGreen}$\uparrow$1.2}} & \shortstack{33.7\\{\scriptsize\color{ForestGreen}$\uparrow$4.4}} \\
\bottomrule
\end{tabular}
}
\label{tab:main}
\end{table*}

\subsection{Evaluation Settings}

\textbf{Training Dataset.}
As described in the Training Data section, we construct a dataset of 2,000 high-quality samples from FineCops-Ref~\cite{liu2024finecops}, balanced with 1,000 positive and 1,000 negative instances. For CRS~\cite{yang2025new}, we follow its official setting, converting the dataset into a multi-choice region-selection format with candidate regions from Grounding DINO~\cite{liu2023grounding}.
In the DPO setting, for positive samples, the model's ``None'' response is treated as the rejected response, while for negative samples, the ground-truth bounding box from the paired positive sample serves as the rejected response.

\begin{figure}[!b]
  \centering
  \includegraphics[width=0.75\linewidth]{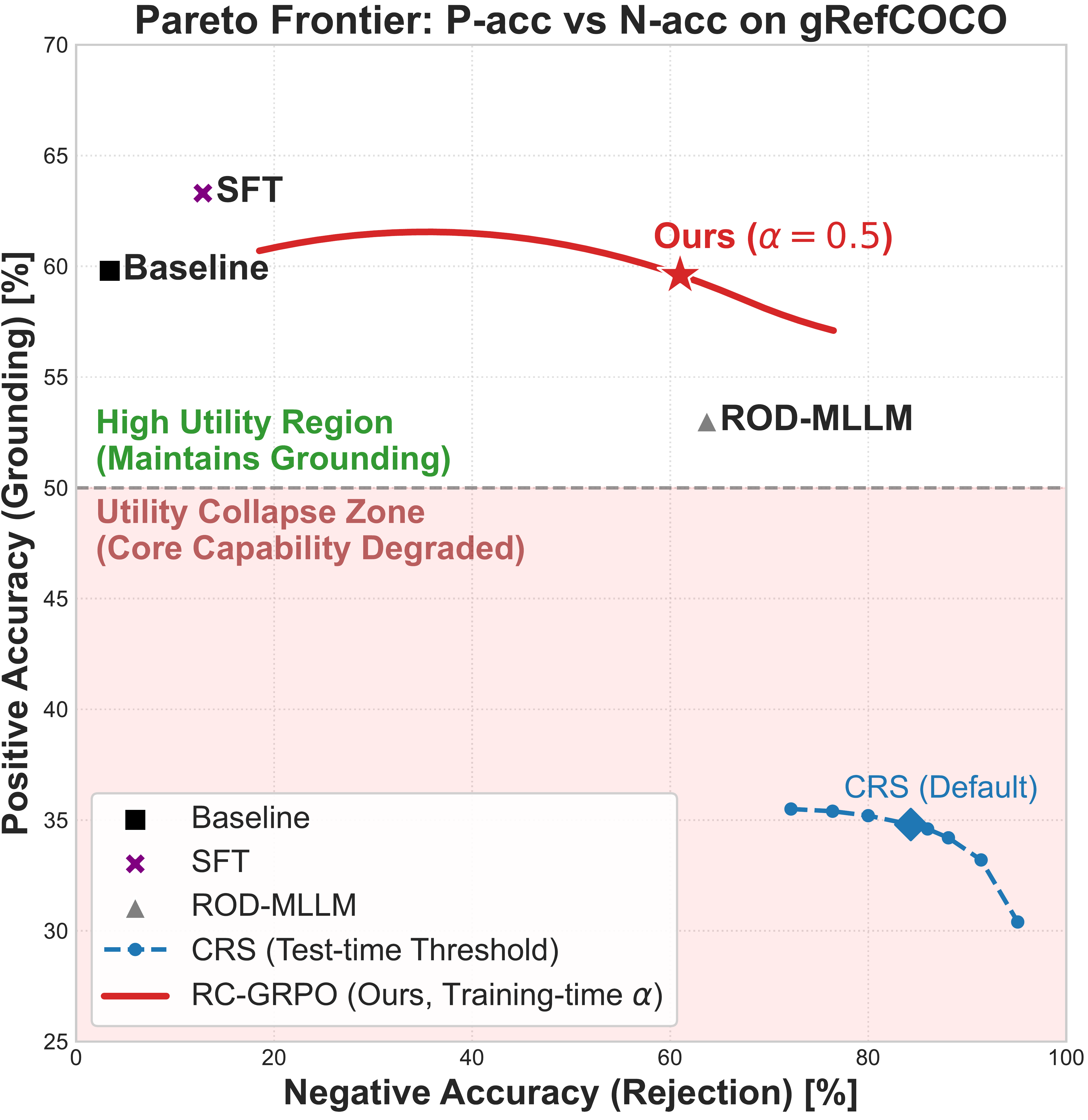}
  \caption{\textbf{Pareto frontier of P-acc vs. N-acc on gRefCOCO.} Baselines like CRS sacrifice grounding accuracy (P-acc) to gain refusal reliability (N-acc), leading to a rapid entry into the utility collapse zone. SFT exhibits poor generalization; despite its high N-acc on FineCops-Ref, its negligible refusal ability here (12.8\%) underscores a failure to learn robust logic. In contrast, RC-GRPO establishes a superior Pareto frontier in the high-utility region.}
  \label{fig:pareto}
\end{figure}

\textbf{Implementation Details.}
% We use LoRA~\cite{hu2022lora} for all training phases. Each phase is trained for one epoch (500 steps) on the 2,000-sample dataset using identical LoRA configurations and learning rates.
% For SFT, training is performed on 2×A800 GPUs with a batch size of 1 and gradient accumulation steps of 2.
% For RL-based fine-tuning, GRPO is implemented under the VLM-R1 framework~\cite{shen2025vlm} and trained on 2 GPUs with 8 sampled generations per input (num\_gens = 8), a batch size of 8, and gradient accumulation steps of 2. The scaling coefficient \( \alpha\ \)for negative advantage scaling is set to 0.5.
% For DPO, we use the ms-swift infrastructure~\cite{zhao2025swift}, training on 2 GPUs under the same gradient settings.
We use LoRA~\cite{hu2022lora} for all training stages, each trained for one epoch (500 steps): SFT on 2×A800 GPUs with batch size 1 and gradient accumulation 2, GRPO with VLM-R1~\cite{shen2025vlm} using 8 generations and batch size 8, and DPO with ms-swift~\cite{zhao2025swift}. More details are provided in the Supplementary Material.

\textbf{Benchmarks and Metrics.}
% We evaluate on three benchmarks that include nonexistent references. FineCops-Ref~\cite{liu2024finecops} is a fine-grained REC benchmark. To align with mainstream GREC setting, we maintain a 3:1 ratio of positive to negative samples, since accurate localization on positive cases remains fundamental.
% gRefCOCO~\cite{liu2023gres} is the most widely used GREC benchmark. For both FineCops-Ref and gRefCOCO, we report Precision@(F1 = 1, IoU $\ge$ 0.5) and nonexistent accuracy (N-acc). 
We evaluate on three benchmarks that include nonexistent references. FineCops-Ref~\cite{liu2024finecops}, a fine-grained REC benchmark, maintains a 3:1 ratio of positive to negative samples, as accurate localization on positive cases remains fundamental.
gRefCOCO~\cite{liu2023gres}, the most widely used GREC benchmark, is evaluated alongside FineCops-Ref using Precision@(F1 = 1, IoU $\ge$ 0.5) and nonexistent accuracy (N-acc). 
% Precision measures the overall performance on both positive and negative samples, defined as the proportion of samples achieving an F1 score of 1 with an IoU threshold of 0.5. N-acc measures the accuracy of rejecting nonexistent references.
Precision (Pr) measures overall accuracy, representing the weighted sum of positive-sample accuracy (P-acc) and N-acc. \textbf{Crucially, a higher Pr indicates a strictly superior model rather than a simple preference trade-off. }
% We emphasize that P-acc represents the foundational grounding capability; a model with P-acc below 50\% is considered functionally deficient for practical deployment. N-acc, in turn, specifically measures the reliability of rejecting nonexistent references.
P-acc measures foundational grounding ability, while N-acc reflects refusal reliability.
D\textsuperscript{3}~\cite{Xie2023DescribedOD} is a Described Object Detection (DOD) benchmark. We report Average Precision (AP), including Presence (Pres), Absence (Abs; e.g., ``refrigerator without fruit''), and the overall metric (Full).

\textbf{Evaluated Methods.}
% We evaluate a range of representative approaches, including both specialist models and Multimodal Large Language Models (MLLMs).
We compare specialist models and Multimodal Large Language Models (MLLMs).
Specialist methods include UNINEXT-R50~\cite{yan2023universal}, HieA2G-R101~\cite{wang2025hierarchical}, and InstanceVG~\cite{dai2025improving}, which are designed specifically for visual grounding.
MLLMs are divided into two categories:
(1) baseline models without explicit nonexistent-rejection training, including Ferret-7B~\cite{you2024ferret}, Qwen2.5-VL-7B-Instruct~\cite{Qwen2.5-VL}, Qwen3-VL-4B-Instruct~\cite{qwen3}, and the larger Qwen3-VL-235B-A22B-Instruct~\cite{qwen3};
(2) nonexistent-aware models, including ROD-MLLM~\cite{yin2025rod} and CRS.
% ROD-MLLM is built on Vicuna-7B v1.5~\cite{vicuna2023} and trained on millions of grounding and GREC samples. CRS adopts a specialist–MLLM collaboration framework for nonexistent references. For fairness, we re-train CRS on our dataset using Qwen2.5-VL-7B-Instruct~\cite{Qwen2.5-VL} and Qwen3-VL-4B-Instruct~\cite{qwen3} as backbones.
ROD-MLLM, built on Vicuna-7B v1.5~\cite{vicuna2023}, is trained on millions of GREC samples, while CRS uses a specialist-MLLM collaboration framework for nonexistent references. For fairness, we re-train CRS on our dataset with Qwen2.5-VL-7B-Instruct~\cite{Qwen2.5-VL} and Qwen3-VL-4B-Instruct~\cite{qwen3} as backbones.

\subsection{Evaluation of the Proposed RC-GRPO}
\textbf{Significant Gains for Models Lacking Rejection Capability.}
% As shown in Table~\ref{tab:main}, our method consistently outperforms existing MLLMs across all three benchmarks. For models such as Qwen2.5-VL-7B-Instruct, which initially lack rejection capability, we observe a 38\% average improvement in Pr (45.0 to 62.2) and 29\% in AP (26.4 to 34.0) compared to the baseline. On gRefCOCO, our approach achieves particularly strong gains, outperforming both ROD-MLLM and CRS by 13\% (53.5 to 60.5) and 24\% (48.8 to 60.5) in Precision, respectively.
% For Qwen3-VL-4B-Instruct, which already possesses a certain level of rejection ability, the advantage of our forced-rejection rollout mechanism is smaller but remains substantial. Compared with the baseline, our method yields an average increase of 3.4 points in Pr and 2.5 points in AP. Notably, on gRefCOCO, it still outperforms ROD-MLLM and CRS by large margins, with Pr improvements of 32\% (53.5 to 70.6) and 41\% (50.1 to 70.6), respectively. More importantly, our method surpasses the current strongest MLLM, Qwen3-VL-235B-A22B-Instruct, on both gRefCOCO and D\textsuperscript{3}, and achieves comparable performance on FineCops-Ref, demonstrating its strong generalization and effectiveness.
As shown in Table~\ref{tab:main}, our method consistently outperforms existing MLLMs across all three benchmarks. For Qwen2.5-VL-7B-Instruct, which initially lacks rejection capability, Pr improves from 45.0 to 62.2 and AP from 26.4 to 34.0; on gRefCOCO, it surpasses ROD-MLLM and CRS by 13\% (53.5→60.5) and 24\% (48.8→60.5) in Precision, respectively.
For Qwen3-VL-4B-Instruct, which already exhibits rejection ability, our method further improves Pr by 3.4 points and AP by 2.5 points; on gRefCOCO, it exceeds ROD-MLLM and CRS by 32\% (53.5→70.6) and 41\% (50.1→70.6) in Precision. It also outperforms Qwen3-VL-235B-A22B-Instruct on both gRefCOCO and D\textsuperscript{3}, while remaining comparable on FineCops-Ref, demonstrating strong generalization and effectiveness.

\textbf{Balanced Rejection Ability with Minimal Degradation on Positive Samples.}
Although CRS attains slightly higher N-acc than our method, it suffers from reduced overall Precision, indicating degraded localization accuracy on positive samples and a higher false-rejection rate. In contrast, RC-GRPO endows MLLMs with robust rejection capability while preserving positive-sample performance, yielding superior overall Precision—particularly on gRefCOCO—and confirming that our approach teaches models when and why to say ``no,'' rather than rejecting all inputs indiscriminately.
%As shown in the table, although CRS attains higher N-acc scores than our method, it performs worse in overall precision. This indicates that CRS tends to harm positive-sample accuracy, leading to an increased false-positive rate. In contrast, our approach enables models to acquire robust nonexistent rejection ability while minimizing the negative impact on positive samples. This balanced trade-off results in superior overall precision, particularly on gRefCOCO. These findings demonstrate that our method effectively teaches the model when and why to say ``no'', rather than merely rejecting all inputs indiscriminately.

\textbf{Pareto Frontier in the High-Utility Region.}
% To further investigate the intrinsic trade-off between localization accuracy and rejection reliability, we illustrate the performance trajectories of different methods in Figure~\ref{fig:pareto}. These curves are generated by varying the test-time rejection thresholds for CRS and the training-time scaling factor $\alpha$ for RC-GRPO. 
% As observed, while CRS can achieve higher N-acc under specific thresholds, it does so at the cost of a sharp collapse in P-acc, quickly falling into the ``utility collapse zone'' where grounding performance becomes practically unusable (P-acc < 50\%). 
% Furthermore, SFT demonstrates a significant lack of generalization: while it shows over-refusal on FineCops-Ref, its rejection accuracy on gRefCOCO drops to a negligible 12.8\% despite maintaining high P-acc (63.3\%). This stark contrast indicates that SFT merely overfits to training distribution frequencies rather than internalizing a generalizable refusal logic.
% In contrast, RC-GRPO establishes a superior \textit{Pareto frontier in the high-utility region}, effectively endowing MLLMs with robust rejection capability while preserving their core localization performance. Our approach teaches models when and why to say ``no,'' rather than simply rejecting all inputs indiscriminately, leading to a more reliable and causally grounded refusal behavior.
Figure~\ref{fig:pareto} illustrates the trade-off between localization accuracy and rejection reliability, obtained by varying CRS's test-time rejection threshold and RC-GRPO's training-time scaling factor $\alpha$. 
While CRS can achieve higher N-acc, it triggers a sharp collapse in P-acc, falling into the ``utility collapse zone'' where grounding performance becomes practically unusable (P-acc < 50\%). 
SFT lacks generalization: despite high P-acc, its N-acc on gRefCOCO drops to 12.8\%, indicating it overfits to training frequencies rather than internalizing refusal logic.
In contrast, RC-GRPO establishes a superior Pareto frontier in the high-utility region, achieving robust rejection while preserving foundational grounding—a more reliable, causally grounded refusal behavior.

\subsection{Comparison with Mainstream Training Strategies}

To further validate the necessity of our framework, we compare RC-GRPO against a broad set of training strategies. We first evaluate mainstream post-training paradigms—Supervised Fine-Tuning (SFT), Group Relative Policy Optimization (GRPO), and Direct Preference Optimization (DPO)—then explore several SFT-based variants to address the optimization imbalance:
\begin{itemize}
\item \textbf{SFT with data balancing}: Applies a 2:1 ratio of positive to negative samples to mitigate the severe rejection bias.
\item \textbf{SFT + GRPO (hybrid loss)}: Combines GRPO loss for positive samples with SFT loss for negative ones via gradient accumulation.
\item \textbf{SFT + Unlikelihood}: Uses unlikelihood training on positive examples to suppress incorrect ``None'' predictions.
\item \textbf{SFT $\to$ GRPO (two-stage)}: A two-stage training pipeline using SFT as a cold-start strategy, which is currently a mainstream approach in post-training.
\end{itemize}

\begin{table}[t]
\centering
\caption{Comparison with mainstream training strategies on FineCops-Ref. 
Pr. denotes overall precision, while P-acc. and N-acc. measure localization and rejection accuracy.}
\resizebox{0.8\columnwidth}{!}{
\begin{tabular}{lccc}
\toprule
\textbf{Training Method} & \textbf{Pr.} & \textbf{P-acc.} & \textbf{N-acc.} \\
\midrule
Qwen2.5-VL-7B (Base) & 56.4 & 73.6 & 5.3 \\
\midrule
SFT (standard) & 46.4 & 30.1 & 95.4 \\
SFT with data balancing & 52.9 & 40.5 & 90.0 \\
SFT + GRPO (hybrid loss) & 43.7 & 26.2 & \textbf{96.0} \\
SFT + Unlikelihood & 64.3 & \textbf{69.8} & 47.6 \\
SFT $\to$ GRPO (two-stage) & 58.2 & 47.7 & 89.7 \\
\midrule
GRPO & 62.5 & 56.9 & 80.0 \\
DPO & 64.3 & 60.9 & 74.8 \\
\midrule
\textbf{RC-GRPO (Ours)} & \textbf{67.4} & 69.6 & 61.1 \\
\bottomrule
\end{tabular}}
\label{tab:training}
\end{table}

As shown in Table~\ref{tab:training}, existing training paradigms suffer from optimization imbalance. Standard SFT achieves an exceptionally high N-acc (95.4\%) but causes positive-sample localization to plummet from 73.6\% to 30.1\%, indicating the model learns indiscriminate rejection over genuine reasoning. The analyzed variants also fail to resolve this intrinsic imbalance: both \textit{SFT with data balancing} and \textit{SFT + GRPO (hybrid loss)} still treat negative samples as ``learning shortcuts,'' causing rapid rejection bias, while \textit{SFT $\to$ GRPO} yields minimal recovery of grounding capability because the policy is initialized from an already SFT-biased state. Although \textit{SFT + Unlikelihood} preserves positive accuracy well (69.8\%), it over-suppresses rejection, resulting in a significantly lower N-acc (47.6\%).

In contrast, RC-GRPO—via contrastive rewards and negative advantage scaling—mitigates this imbalance, maintaining high positive accuracy while steadily improving nonexistent accuracy along an interpretable trajectory (training dynamics in the Supplementary Material). This balanced evolution enables reliable rejection without compromising localization, ensuring the model learns when and why to refuse—a key step toward trustworthy GREC.

%Figure~\ref{fig:training} visually demonstrates this result. Both SFT and baseline GRPO gradually shift their optimization focus towards negative examples, which led to the model achieving higher accuracy on negative samples than on positive samples, significantly impacting the model's ability to accurately locate positive cases. In contrast, our approach maintains a high level of positive accuracy while progressively improving nonexistent accuracy, effectively striking a balance.
%This enables the model to acquire the rejection capability while minimizing the degradation of positive-sample performance, resulting in a more balanced and interpretable optimization outcome. 

% \begin{table}[h]
% \centering
% \caption{\textbf{Ablation on Key Components in RC-GRPO-I on FineCops-Ref.} 
% Each ablation removes a specific module from RC-GRPO-I while keeping all other components unchanged.}
% \resizebox{0.48\textwidth}{!}{
% \begin{tabular}{lccc}
% \toprule
% \textbf{Method} & \textbf{Pr.} & \textbf{P-acc.} & \textbf{N-acc.} \\
% \midrule
% \multicolumn{4}{l}{\textit{\textbf{Baseline}}} \\
% Qwen2.5-VL-7B-GRPO & 62.5 & 56.9 & 80.0 \\
% \midrule
% \multicolumn{4}{l}{\textit{\textbf{Ablation}}} \\
% w/o Forced-Rejection Rollout & 65.5 & 68.0 & 58.4 \\
% w/o Contrastive Reward & 65.0 & 63.5 & 69.9 \\
% w/o Negative Advantage Scaling & 64.3 & 60.9 & \textbf{74.5} \\
% \midrule
% \multicolumn{4}{l}{\textit{\textbf{Ours}}} \\
% RC-GRPO-I & \textbf{66.6} & \textbf{69.0} & 59.5 \\
% \bottomrule
% \end{tabular}}
% \label{tab:ablation}
% \end{table}

\subsection{Ablation Study}
In this section, we conduct experiments on FineCops-Ref to evaluate the contribution of individual components and the sensitivity of the scaling factor $\alpha$.

\begin{table}[h]
\centering
\caption{Ablation on key components in RC-GRPO-I on FineCops-Ref.
Each ablation removes a specific module from RC-GRPO-I while keeping all other components unchanged.}
\resizebox{0.8\columnwidth}{!}{
\begin{tabular}{lccc}
\toprule
\textbf{Method} & \textbf{Pr.} & \textbf{P-acc.} & \textbf{N-acc.} \\
\midrule
Qwen2.5-VL-7B-GRPO & 62.5 & 56.9 & 80.0 \\
\midrule
w/o Forced-Rejection Rollout & 65.5 & 68.0 & 58.4 \\
w/o Contrastive Reward & 65.0 & 63.5 & 69.9 \\
w/o Negative Advantage Scaling & 64.3 & 60.9 & \textbf{74.5} \\
\midrule
\textbf{RC-GRPO-I} & \textbf{66.6} & \textbf{69.0} & 59.5 \\
\bottomrule
\end{tabular}}
\label{tab:ablation_components}
\end{table}

\subsubsection{Component-Wise Ablation Study}
% Table~\ref{tab:ablation} presents an ablation study evaluating the contribution of the three key components in the RC-GRPO-I training framework: forced-rejection rollout, contrastive reward, and negative advantage scaling. The results show that each component plays a crucial and complementary role in achieving balanced optimization.

Table~\ref{tab:ablation_components} evaluates the three key components of RC-GRPO-I: forced-rejection rollout, contrastive reward, and negative advantage scaling. The results demonstrate that each module plays a complementary role in achieving balanced optimization. 

Specifically, the forced-rejection rollout is essential for early-stage exploration; without it, the model cannot compute valid advantages for negatives, hindering stable learning. The contrastive reward provides consistent optimization signals for both positive and negative cases, stabilizing training. Negative advantage scaling acts as a critical balancer: removing it causes the model to overfit the simpler rejection task, sharply dropping positive-sample accuracy. Together, these components ensure reliable rejection without sacrificing core localization competence.
% Specifically, forced-rejection rollout enhances exploration in the early training stages by ensuring valid advantages for negative samples, allowing the model to effectively learn the ``None'' behavior. When combined with contrastive reward, the model receives consistent optimization signals for both positive and negative cases, leading to a more stable training process. Furthermore, negative advantage scaling mitigates the inherent imbalance between easy negative and difficult positive samples, preventing overfitting to rejection.

% The three components jointly contribute to the success of RC-GRPO-I, ensuring that the model develops reliable rejection capability without sacrificing localization accuracy.
%In Table~\ref{tab:ablation}, we present an ablation study to evaluate the contribution of three key components in the RC-GRPO-I training framework: forced-rejection rollout, contrastive reward, and negative advantage scaling. The results indicate that each of these components contributes positively to the overall performance. Specifically, forced-rejection rollout enhances the model's exploration during the early stages of training. When combined with the contrastive reward, it provides the model with a stable direction for optimization. Additionally, both the contrastive reward and negative advantage scaling help address the optimization imbalance between positive and negative samples. Collectively, these three components are essential to the successful training of the model.

\begin{table}[t]
  \centering
  \caption{\textbf{Sensitivity analysis of the scaling factor $\alpha$ on FineCops-Ref}. Results show that $\alpha=0.5$ achieves the optimal balance between localization and rejection.}
  \resizebox{0.7\columnwidth}{!}{
  \begin{tabular}{cccc}
    \toprule
    \textbf{Scaling Factor $\alpha$} & \textbf{Pr.} & \textbf{P-acc.} & \textbf{N-acc.} \\
    \midrule
    0.2 & 60.1 & \textbf{73.6} & 19.6 \\
    0.5 & \textbf{66.6} & 69.0 & 59.5 \\
    0.8 & 64.7 & 62.7 & 70.8 \\
    1.0 & 64.3 & 60.9 & \textbf{74.5} \\
    \bottomrule
  \end{tabular}}
  \label{tab:alpha_sensitivity}
\end{table}

\subsubsection{Sensitivity of Scaling Factor $\alpha$}
We further investigate the negative advantage scaling factor $\alpha$, which controls the optimization balance between negative and positive samples, as presented in Table~\ref{tab:alpha_sensitivity}. 
A low value ($\alpha=0.2$) overly limits the gradient signal from negative samples, so the model struggles to acquire refusal behavior, as evidenced by a low N-acc of 19.6\%. As $\alpha$ increases toward 1.0, the optimization biases heavily toward the simpler rejection task, progressively degrading localization accuracy (P-acc drops from 73.6\% to 60.9\%). We find that $\alpha=0.5$ yields the optimal balance, achieving the highest overall Precision (66.6) by maintaining robust grounding while establishing reliable rejection logic.

\subsection{Generalization to General Multimodal Tasks}

\begin{table}[h]
\centering
\caption{Performance on general multimodal and hallucination benchmarks. Results indicate that RC-GRPO incurs negligible degradation on general capabilities and maintains robust anti-hallucination performance.}
\resizebox{0.7\columnwidth}{!}{
\begin{tabular}{lccc}
\toprule
\textbf{Method} & \textbf{MME} & \textbf{MMBench} & \textbf{POPE} \\
\midrule
Baseline & 2304.5 & 82.6 & 83.6 \\
RC-GRPO-I & 2282.7 & 82.2 & 83.6 \\
RC-GRPO-II & 2294.2 & 82.5 & 83.7 \\
\bottomrule
\end{tabular}}
\label{tab:general}
\end{table}

To assess the impact of RC-GRPO on the base model's foundational performance, we evaluate on three mainstream benchmarks: MME~\cite{fu2023mme}, MMBench~\cite{liu2024mmbench}, and POPE~\cite{li2023evaluating}.
As presented in Table~\ref{tab:general}, RC-GRPO remains highly stable across all metrics: on MME and MMBench, our models score nearly identical to the baseline, while on POPE, which measures object hallucination, RC-GRPO-II even yields a marginal improvement. These results confirm that our framework adds robust refusal without compromising general multimodal understanding or inducing additional hallucinations.

\FloatBarrier

\section{Conclusion}
\label{sec:conclusion}

% In this work, we address the underexplored Generalized Referring Expression Comprehension (GREC) task, which requires both precise localization of existing objects and reliable rejection of nonexistent ones. We propose Refusal-Calibrated Group Relative Policy Optimization (RC-GRPO), a reinforcement learning framework that calibrates refusal behavior while preserving localization accuracy. A reasoning-enhanced stage further reinforces causal understanding of object absence. Experiments on multiple benchmarks show that RC-GRPO achieves a better balance between localization and refusal.

In this work, we address the underexplored Generalized Referring Expression Comprehension (GREC) task, which requires both precise localization of existing objects and reliable rejection of nonexistent ones. To overcome existing MLLMs' tendency to hallucinate bounding boxes, we propose Refusal-Calibrated Group Relative Policy Optimization (RC-GRPO), a post-training framework that calibrates refusal behavior while preserving localization accuracy via refusal calibration and a positive-sample penalty; a reasoning-enhanced stage further reinforces causal understanding of object absence. Experiments on multiple GREC benchmarks show that RC-GRPO achieves a superior balance between localization accuracy and refusal reliability, advancing trustworthy multimodal reasoning.
%In this work, we tackle the underexplored Generalized Referring Expression Comprehension (GREC) task, which requires models to both accurately localize existing objects and reliably reject nonexistent ones. To overcome the limitations of existing MLLMs that tend to hallucinate bounding boxes, we propose Refusal-Calibrated Group Relative Policy Optimization (RC-GRPO), a reinforcement learning–based post-training framework that jointly optimizes localization and refusal behaviors through refusal calibration and positive-sample penalty mechanisms. Furthermore, a reasoning-enhanced reinforcement stage strengthens the model’s causal understanding of object absence. Experiments on multiple GREC benchmarks demonstrate that RC-GRPO achieves a superior balance between localization accuracy and refusal reliability, advancing the development of trustworthy multimodal reasoning systems.

\section{Additional Analyses and Experiments}
\label{sec:additional}

This appendix consolidates additional analyses and supplementary experiments that further validate the effectiveness, robustness, and generality of RC-GRPO.

\subsection{On the Training Bias of Forced Rejection}
\begin{figure}[h]
    \centering
    \includegraphics[width=0.8\columnwidth]{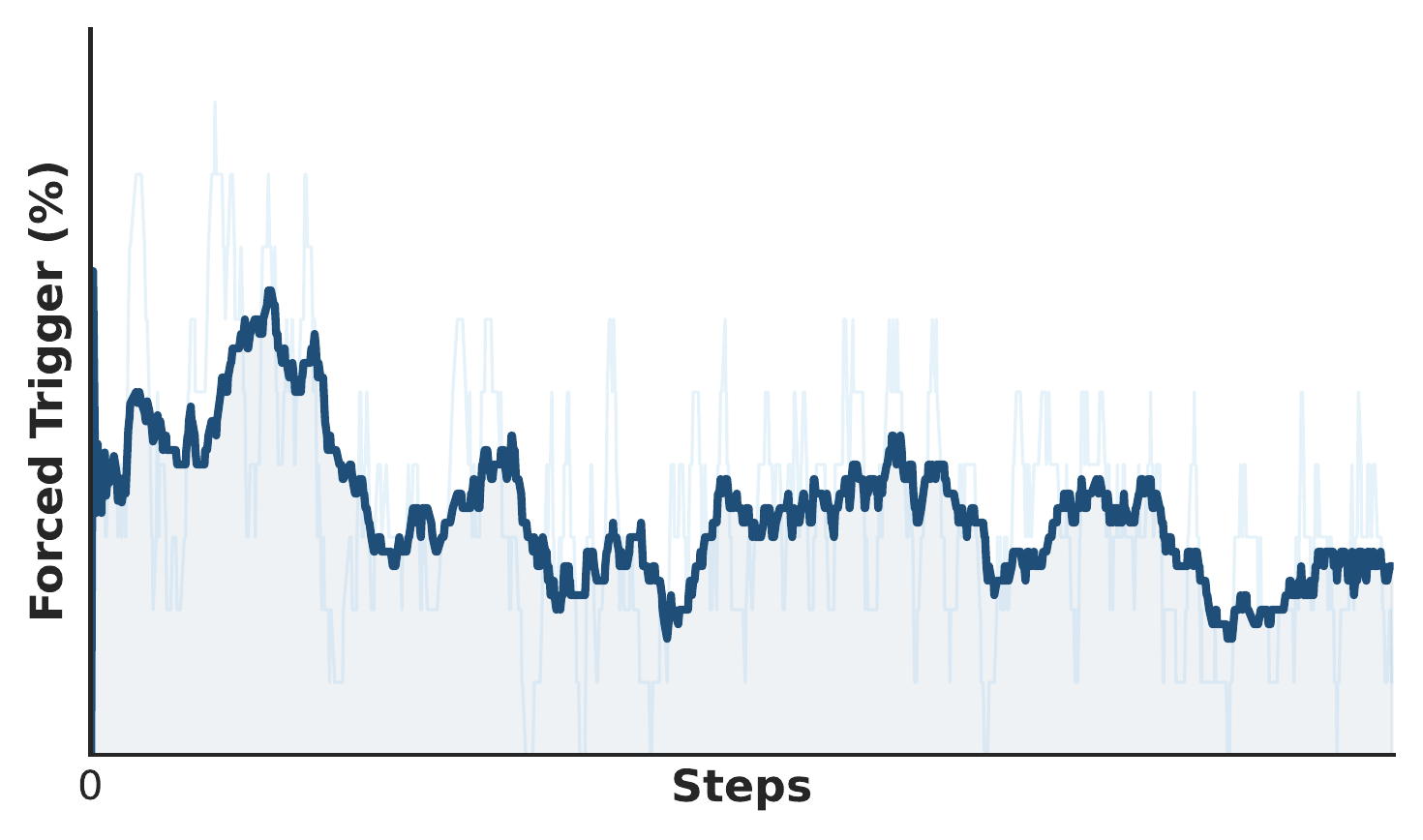}
    \caption{The forced-rejection trigger rate decays naturally as training progresses.}
    \label{fig:trigger}
\end{figure}

Unaligned models rarely emit ``None'' and default to hallucinated boxes, collapsing advantage estimation. While forced-rejection indeed introduces an off-policy bias by upsampling a high-value but low-probability path, the intervention rate decays significantly as the policy internalizes refusal logic (Figure~\ref{fig:trigger}). Training thus reverts to predominantly on-policy updates in its mid-to-late stages. General benchmarks confirm that this residual bias leaves the model's foundational capabilities intact. For forced outputs, we calculate the sequence probability $\pi_\theta(\text{``None''} \mid x)$ using current logits, assign rewards, and estimate advantages exactly like standard trajectories.

\subsection{Training Dynamics of RC-GRPO}
\begin{figure}[ht]
  \centering
  \includegraphics[width=0.9\columnwidth]{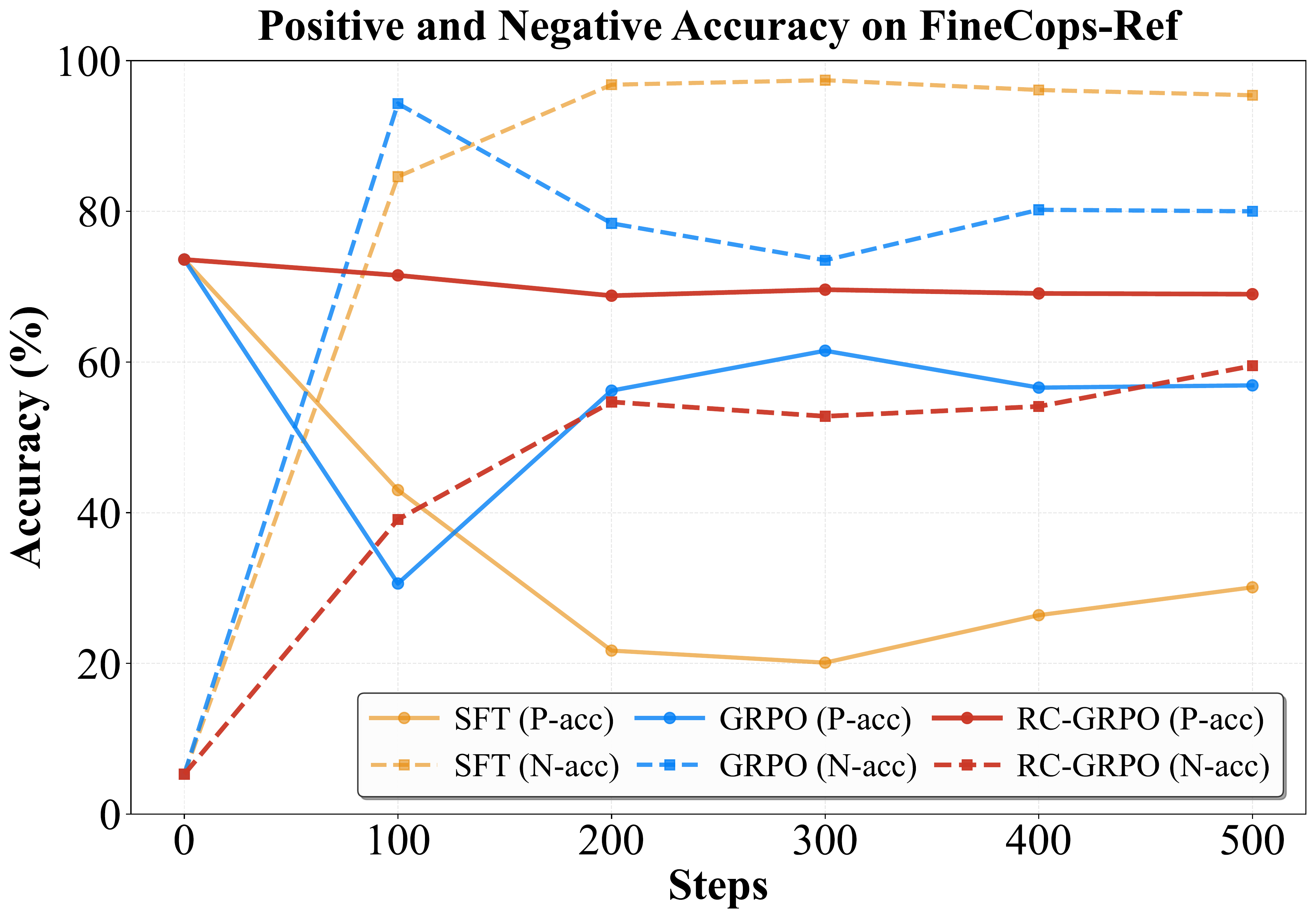}
  \caption{Accuracy changes of SFT, GRPO, and RC-GRPO-I over training steps. The yellow, blue, and red curves denote SFT, baseline GRPO, and RC-GRPO-I, respectively. Solid lines show P-acc., and dashed lines show N-acc. SFT and GRPO overfit to negatives—achieving higher accuracy on nonexistent cases than positives—whereas RC-GRPO-I maintains balanced performance across both.}
  \label{fig:training}
\end{figure}

As illustrated in Figure~\ref{fig:training}, both SFT and GRPO gradually shift their optimization toward negative samples, eventually producing higher accuracy on negatives than on positives, which severely harms the model's grounding capacity. RC-GRPO, however, maintains high positive accuracy while steadily improving nonexistent accuracy, achieving an interpretable optimization trajectory. This balanced evolution confirms that the contrastive reward and negative advantage scaling jointly enable reliable rejection without compromising localization.

\subsection{Rationale Evaluation}
To dispel concerns of ``template hacking'' (memorizing ``not X but Y''), we conducted evaluations via GPT-4o and human on 200 refusal cases. We assess \textbf{Faithfulness} (\textbf{Fa.}, alignment with visual facts), \textbf{Logical Validity} (\textbf{LV}, soundness of causal reasoning), and \textbf{Linguistic Richness} (\textbf{LR}, phrasing diversity). RC-GRPO achieves consistent improvements (Table~\ref{tab:rationale}). Moreover, our zero-shot gains on natively non-template benchmarks (gRefCOCO and $D^3$) prove that RC-GRPO acquires genuine causal rejection.

\begin{table}[h]
    \centering
    \caption{Rationale evaluation on a 5-point scale (GPT-4o and human raters).}
    \resizebox{\columnwidth}{!}{
    \begin{tabular}{l|ccc|ccc}
    \toprule
    & \multicolumn{3}{c|}{\textit{GPT-4o}} & \multicolumn{3}{c}{\textit{Human}} \\
    \cmidrule(lr){2-4}\cmidrule(lr){5-7}
    \textbf{Method} & \textbf{Fa.}$\uparrow$ & \textbf{LV}$\uparrow$ & \textbf{LR}$\uparrow$ & \textbf{Fa.}$\uparrow$ & \textbf{LV}$\uparrow$ & \textbf{LR}$\uparrow$ \\
    \midrule
    SFT & 2.56 & 2.39 & 2.19 & 2.90 & 2.71 & 2.57 \\
    \textbf{Ours} & \textbf{3.64} & \textbf{4.03} & \textbf{3.04} & \textbf{3.71} & \textbf{3.79} & \textbf{3.15} \\
    \bottomrule
    \end{tabular}}
    \label{tab:rationale}
\end{table}

\subsection{Sampling, Distribution, and Data Scaling}
We clarify ``high-quality'' merely means excluding corrupted annotations. The 2K subset is drawn by stratified random sampling, preserving the original difficulty \textit{levels} and mismatch \textit{types}: Positive (Level-1 64.4\%, Level-2 35.6\%) and Negative (Object 36.1\%, Attribute 26.3\%, Relation 20.2\%, Order 11.3\%, Swap 6.1\%). To prove robustness, we evaluated a newly re-sampled 2K set, yielding \textbf{66.5\% Pr}, nearly identical to our reported 66.6\%. Table~\ref{tab:scaling} shows the scaling trend, where performance efficiently \textbf{converges at $N=2K$}. These prove our sample-efficiency stems from the RL framework, not curated data.

\begin{table}[h]
    \centering
    \caption{Effect of training-set size on performance.}
    \resizebox{0.8\columnwidth}{!}{
    \begin{tabular}{c|ccc}
    \toprule
    \textbf{Size ($N$)} & \textbf{Pr.} & \textbf{P-acc.} & \textbf{N-acc.} \\
    \midrule
    0.5K & 63.0 & 58.7 & 75.9 \\
    1K & 65.6 & 67.6 & 59.5 \\
    \textbf{2K (Ours)} & \textbf{66.6} & \textbf{69.0} & \textbf{59.5} \\
    5K & 65.4 & 64.2 & 69.5 \\
    \bottomrule
    \end{tabular}}
    \label{tab:scaling}
\end{table}

\subsection{A Stronger DPO Baseline}
Following guidance, we constructed a stronger DPO baseline using the base model's actual misaligned boxes as rejected responses. Empirically, it suffered a \textbf{Catastrophic Coordinate Collapse}: P-acc plummeted to \textbf{0.6\%} (N-acc: 65.8\%). We hypothesize that penalizing boxes in both branches lets the model reward-hack by suppressing all box tokens to minimize loss, thereby erasing its grounding capability. This failure exposes the intrinsic optimization imbalance of existing methods. Conversely, RC-GRPO internalizes refusal logic with minimal degradation to foundational localization.

\subsection{Generalization to Other Backbones}
Validating generalizability on MiniCPM-V (Table~\ref{tab:minicpm}), RC-GRPO consistently achieves a superior balance, boosting Pr to 53.0\% and avoiding the severe P-acc collapse in SFT.

\begin{table}[h]
    \centering
    \caption{Results on MiniCPM-V-4.6.}
    \resizebox{0.8\columnwidth}{!}{
    \begin{tabular}{l|ccc}
    \toprule
    \textbf{Method} & \textbf{Pr.} & \textbf{P-acc.} & \textbf{N-acc.} \\
    \midrule
    Base & 46.6 & 61.8 & 1.0 \\
    SFT & 47.2 & 36.1 & 80.5 \\
    \textbf{Ours} & \textbf{53.0} & \textbf{52.8} & \textbf{53.8} \\
    \bottomrule
    \end{tabular}}
    \label{tab:minicpm}
\end{table}

\subsection{Parameter Robustness}
\begin{figure}[h]
    \centering
    \includegraphics[width=0.9\columnwidth]{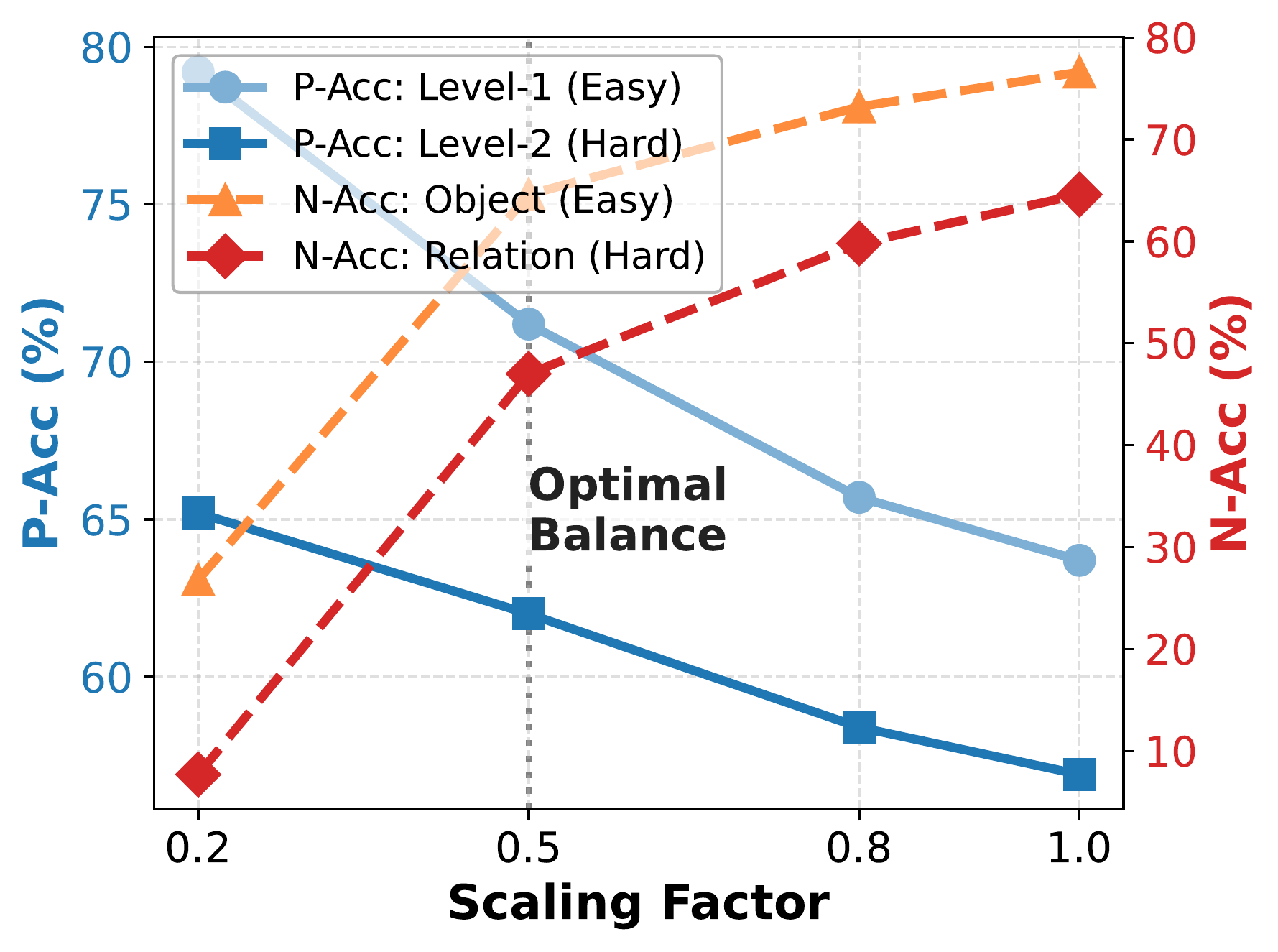}
    \caption{Trend of P-acc and N-acc as the scaling factor $\alpha$ varies.}
    \label{fig:trend}
\end{figure}

The scaling factor $\alpha$ predictably balances grounding and refusal. To demonstrate stability, we analyzed its trend across task complexities (Figure~\ref{fig:trend}). As $\alpha$ increases, grounding (P-acc) drops monotonically while refusal (N-acc) strictly improves. Crucially, $\alpha=0.5$ is a \textbf{Universal Safe Default}: it averts the P-acc collapse at higher values while still activating reasoning on hard negatives---a robust trade-off needing no per-dataset tuning.

\subsection{Additional Clarifications}
We resolve GREC's inherent optimization imbalance problem. Standard GRPO fails in early exploration as models rarely sample ``None'' (advantage vanishes), while SFT warm-ups severely degrade positive localization. Our RC-GRPO overcomes these bottlenecks, significantly \textbf{boosting Pr from 62.5\% (GRPO) to 67.4\%}. On gRefCOCO (val), RC-GRPO (Qwen2.5-7B) lifts Pr to \textbf{64.9\%}, far above base (38.2\%), ROD-MLLM (53.2\%), and CRS (54.9\%). On FineCops-Ref, under identical settings, negative-sample SFT collapses to 46.4\% Pr, whereas RC-GRPO reaches \textbf{67.4\%}, confirming RC-GRPO's clear edge over SFT.

\section{Qualitative Results}
\begin{figure*}[h]
  \centering
  \includegraphics[width=0.95\textwidth]{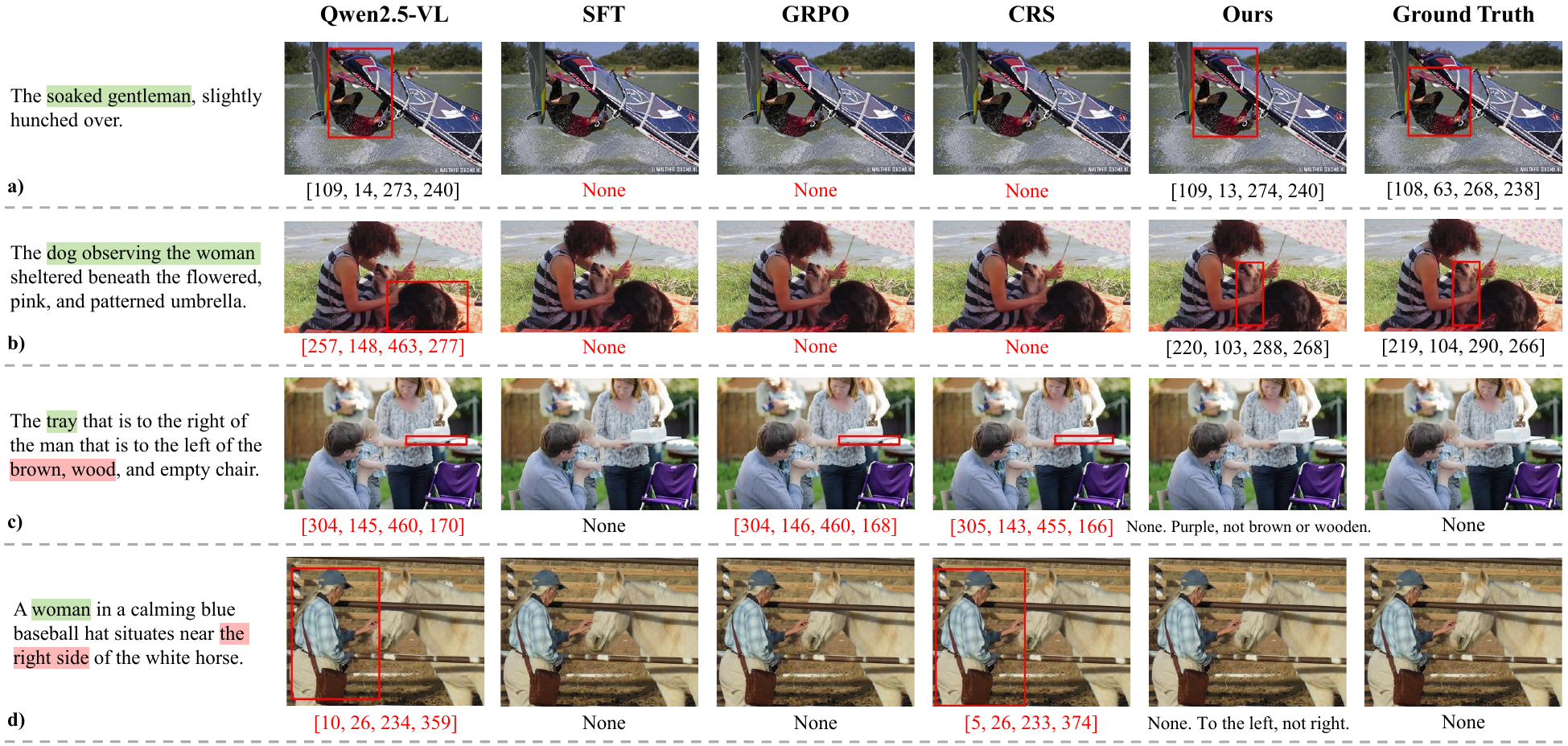}
  \caption{\textbf{Qualitative results on the GREC task.} Compared methods include Qwen2.5-VL~\cite{Qwen2.5-VL}, SFT, GRPO, CRS~\cite{yang2025new}, and our RC-GRPO. In each referring expression, green text marks the main target and red text highlights invalid referring cues. Predicted outputs shown in red denote incorrect predictions.}
  \label{fig:qualitative}
\end{figure*}

Figure~\ref{fig:qualitative} presents the qualitative results. MLLMs such as Qwen2.5-VL consistently predict bounding boxes regardless of the underlying semantics, while models trained with SFT tend to overfit and indiscriminately output ``None.'' In the positive case (b), which requires complex semantic comprehension and precise localization, our method accurately locates the dog observing the woman. In the negative cases (c) and (d) that involve subtle mismatches in attributes or relations, our method effectively detects the inconsistency between the expression and the image, correctly outputs ``None,'' and provides precise, interpretable explanations for the rejection. These examples demonstrate that our approach achieves superior localization capabilities and deep semantic understanding compared to existing methods, ensuring reliable visual grounding for real-world applications.

% The aaai2027.sty file already sets the bibliography style (aaai2027.bst),
% so no \bibliographystyle command is used here.
\bibliography{main}

% The technical appendix (dataset details + additional analyses) has been
% split into a standalone supplementary document (supplementary.tex) so the
% main paper stays within the AAAI page limits.

\end{document}